\documentclass{SCIS2025}
\usepackage{amsmath}
\usepackage{amssymb}
\usepackage{graphicx}
\usepackage[numbers]{natbib}
\begin{document}
\ArticleType{RESEARCH PAPER (PrePrint, Under Review)}
\Year{2025}
\Month{January}
\Vol{68}
\No{1}
\DOI{}
\ArtNo{}
\ReceiveDate{}
\ReviseDate{}
\AcceptDate{}
\OnlineDate{}
\AuthorMark{}
\AuthorCitation{}

\title{Is Compression Really Linear with \\ Code Intelligence?}{Title for citation}


\author[1,3\dag]{Shijie Xuyang}{}
\author[2\dag]{Xianzhen Luo}{}
\author[2]{Zheng Chu}{}
\author[1,3]{Houyi Li}{}
\author[1,3]{Siming Huang}{}
\author[3]{\\Qiufeng Wang}{}
\author[2]{Wanxiang Che}{}
\author[2]{Qingfu Zhu}{{qfzhu@ir.hit.edu.cn}}
\author[1]{Shuigeng Zhou}{{sgzhou@fudan.edu.cn}}


\address[1]{Fudan University, Shanghai 200433, China}
\address[2]{Harbin Institute of Technology, Harbin 150001, China}
\address[3]{StepFun, Beijing 100080, China}

\abstract{Understanding the relationship between data compression and the capabilities of Large Language Models (LLMs) is crucial, especially in specialized domains like code intelligence. 
Prior work posited a linear relationship between compression and general intelligence. However, it overlooked the multifaceted nature of code that encompasses diverse programming languages and tasks, and struggled with fair evaluation of modern Code LLMs.
We address this by evaluating a diverse array of open-source Code LLMs on comprehensive multi-language, multi-task code benchmarks.
To address the challenge of efficient and fair evaluation of pre-trained LLMs' code intelligence, we introduce \textit{Format Annealing}, a lightweight, transparent training methodology designed to assess the intrinsic capabilities of these pre-trained models equitably. 
Compression efficacy, measured as bits-per-character (BPC), is determined using a novel, large-scale, and previously unseen code validation set derived from GitHub. 
Our empirical results reveal a fundamental logarithmic relationship between measured code intelligence and BPC. 
This finding refines prior hypotheses of linearity, which we suggest are likely observations of the logarithmic curve's tail under specific, limited conditions. 
Our work provides a more nuanced understanding of compression's role in developing code intelligence and contributes a robust evaluation framework in the code domain. }

\keywords{large language model, code compression, bit per character, benchmark, scaling law}

\maketitle

\section{Introduction}

Data compression seeks to represent information using fewer bits than its original encoding by exploiting statistical redundancies~\cite{bookstein1990compression, johnson2003introduction, rissanen2000coding,yvinec2021red}. 
According to Kolmogorov complexity theory~\cite{ruffini2007information, li2008introduction,yoran2025the}, effective compression relies upon identifying latent regularities and structural features within data, a process fundamental to intelligence and its core cognitive abilities such as learning, comprehension, and generalization~\cite{zanto2011causal, bargmann2014brain}.
Recently, the next-token prediction objective, central to the pre-training of LLMs, has been recognized as mirroring compression principles~\cite{deletang2024language}. 
This has led to viewing LLMs as powerful compressors and exploring compression-based metrics for evaluating their learned capabilities~\cite{deletang2024language,wang2024language,huang2025look,huang2024compression}.
However, code as a unique data modality differs from natural language in its formal syntactic structures and abstraction mechanisms~\cite{294586,zhu-etal-2024-survey,parțachi2024bringing}. 
For instance, information missing in code can have far more severe consequences~\cite{aryabumi2024code}.
Therefore, a dedicated investigation into code compression and code intelligence is of paramount importance~\cite{gnvv2023the}.

\begin{figure*}
\begin{center}
   \includegraphics[width=1\linewidth]{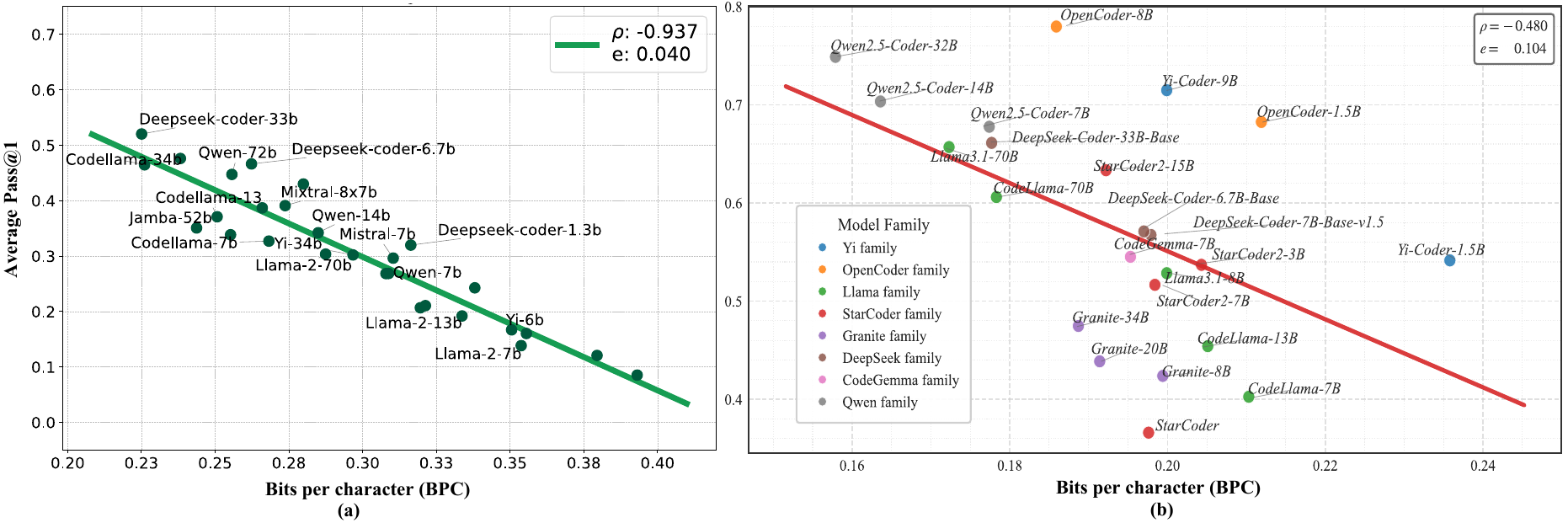}
\end{center}
   \caption{(a) shows the linear relationship between code BPC and code intelligence as claimed in~\cite{huang2024compression}. (b) illustrates the linear fitting result of advanced Code LLMs on the same setting.}
\label{fig:baseline}
\end{figure*}

The previous study~\cite{huang2024compression} proposes the linear relationship between code compression and code intelligence, using bits per character (BPC) as a proxy for the former and average performance on three Python code completion benchmarks: HumanEval~\cite{chen2021evaluating}, MBPP~\cite{austin2021program}, and DS-1000~\cite{lai2023ds}.
Furthermore, evaluated LLMs are earlier versions, which possessed relatively weaker code capabilities and had seen limited adoption in current applications.
Our preliminary evaluation result illustrated in Figure~\ref{fig:baseline}(b) reveals no clear correlation between the average performance of several modern code LLMs and their code BPC.
Code intelligence spans numerous programming languages (e.g., Python, C++, and Java) and encompasses several fundamental ``atomic'' tasks - code understanding (explain), reasoning, generation, and repair~\cite{chen2021evaluating}.
Consequently, evaluating code intelligence solely through Python code completion is insufficiently comprehensive. 
Furthermore, code intelligence has seen significant progress with the emergence of advanced Code LLMs. These LLMs are pre-trained on substantially larger code corpora, exhibit superior performance across a variety of coding tasks, and are increasingly utilized in real-world development environments.
This necessitates a more in-depth study of the relationship between code compression and code intelligence.

Even with a well-defined set of tasks and models, evaluating the code intelligence of pretrained LLMs remains a challenge. 
They often struggle with tasks beyond simple code completion. For instance, they may generate code with non-standard formatting or misinterpret task requirements~\cite{raychev2014code}. 
Common evaluation paradigms like few-shot prompting~\cite{reynolds2021prompt,ye2022unreliability} heavily rely on the choice and quality of demonstrations, exhibit limited generalization, and require significant effort in designing specific prompts for each distinct task. 
Furthermore, some LLMs incorporate instruction-tuning data during their pre-training phase, which artificially inflates performance on benchmarks. 
Illustratively, Figure~\ref{fig:baseline}(b) highlights inconsistencies in the self-reported performance of various Code LLMs. 
These factors highlight an urgent need for more robust and fair evaluation methods capable of revealing the genuine capabilities of pre-trained Code LLMs.


To address these pressing issues, we undertake a two-pronged approach. Firstly, to establish a comprehensive scope for code intelligence, we have curated a diverse suite of open-source Coder LLMs, encompassing prominent families such as Qwen Coder~\cite{hui2024qwen2}, Code Llama~\cite{roziere2023code}, Starcoder~\cite{li2023starcoder} and Opencoder~\cite{huang2024opencoder}. 
Furthermore, we propose OmniCodeBench, which spans multiple programming languages and diverse code-related tasks.
Secondly, to ensure a fair and veridical assessment of pre-trained LLMs, we introduce Format Annealing, which is a lightweight, controlled training method on a consistent and publicly available dataset. 
This process is designed to elicit the intrinsic ability of each model in a standardized manner, reflecting their true intelligence. 
Subsequently, to quantify compression efficacy, we have meticulously constructed a large-scale, high-quality code validation set derived from GitHub, ensuring it is previously unseen by the models, diverse in its content, and unbiased in its composition. 
Using this validation set, we measure the bits-per-character (BPC) of each LLM. 
The empirical results reveal a significant departure from previous findings: 
\textbf{Code Intelligence exhibits a logarithmic relationship with Code BPC}.
We posit that earlier observations of linearity likely stemmed from analyses focused on higher compression regimes, where the tail of a logarithmic curve can indeed approximate linearity.

Our contributions are summarized as below:
    1) We establish a robust and comprehensive framework for investigating the relationship between code intelligence and compression by selecting high-quality, multi-task, and multi-language code benchmarks to construct OmniCodeBench, curating a diverse set of contemporary open-source Coder LLMs, and constructing a novel, large-scale code validation set.
    2) We introduce \textit{Format Annealing}, a principled and transparent methodology for eliciting and fairly evaluating the performance of pre-trained models on these benchmarks using publicly accessible data.
    3) Our empirical investigation reveals a fundamental \textbf{logarithmic} relationship between code compression rate and Code Intelligence, offering a new perspective that refines and contextualizes previous findings of linearity.

\section{Background \& Related Work}
\subsection{Compression as Metric}
Early evaluations of LLM intelligence primarily relied on averaged performance across various benchmarks~\cite{brown2020language}.
However, the scope of benchmarks was constrained by the pretrained LLMs' nascent instruction-following capabilities. 
Recognizing the intrinsic consistency between pre-training objectives and information compression, researchers began to investigate the relationship between model compression rates and intelligence levels.
\citet{wei2023skywork} first reveals the linear correlation between compression ratios and downstream benchmark metrics.
Subsequent studies~\cite{huang2024compression} further substantiated the persistence of this linearity across diverse model series and benchmark domains.
Despite these advancements, prior work has predominantly treated code as a subdomain of natural language, which overlooks fundamental multi-dimensional distinctions. 
Code adheres to strict, formal syntactic rules, precluding the ambiguities and flexible grammatical structures inherent in natural language~\cite{zhu-etal-2024-survey}. 
Furthermore, abstraction in code manifests through explicit, formally-defined constructs such as classes, functions, and APIs, contrasting sharply with the predominantly cognitive and conceptual nature of abstraction in natural language~\cite{de2003side,abbott2008abstraction}. 
These critical differences motivate a dedicated inquiry into whether the observed relationship between compression and intelligence in natural language extends uniformly to code.

\subsection{Code Intelligence}

The power of LLMs in code-related tasks has spurred the concept of ``code intelligence''—the comprehensive ability to understand, generate, and manipulate code~\cite{rabin2021understanding,wan2024deep}.
Early evaluations typically focused on Python code completion tasks, such as HumanEval and MBPP~\cite{zhu-etal-2024-survey}.
However, as code applications and workflows grow in complexity~\cite{wang2024agents,roberttransforming}, it is increasingly evident that a robust notion of code intelligence must extend beyond unilingual assessments to encompass proficiency across multiple programming languages~\cite{cassano2022multipl,peng-etal-2024-humaneval,chai2024mcevalmassivelymultilingualcode,liu2024mdeval}.
Furthermore, a growing body of research suggests that comprehensive code intelligence is not a monolithic entity but 
consists of a set of fundamental or ``atomic'' tasks. 
While existing taxonomies for code-related LLM tasks vary—for instance, by input/output modalities (e.g., NL-PL, PL-NL)  or by user intent~\cite{chen2024survey} —they converge on a core set of underlying capabilities. 
We classify them to 
(1) Code Generation: automated creation of source code from natural language descriptions~\cite{chen2021evaluating,austin2021program,zhuo2024bigcodebench}.
(2) Code Explanation (or Understanding): grasp and articulate the syntax, semantics, functionality, and underlying intent of a given code snippet~\cite{meyerson2025position,fang2024large,nam2024using}.
(3) Code Reasoning: infer code properties and predict code behavior without execution~\cite{dehghan2024assessing,chen2024reasoning}.
(4) Code Repair: identify and generate patches to correct bugs, vulnerabilities, or other errors in source code~\cite{pasquale2025challenges,liu2024mdeval}.
While ``code intelligence'' is a broad concept, its decomposition into the 4 atomic tasks provides a more structured and actionable framework for quantitative evaluation.


\section{Code Compression as Code Intelligence}
\label{sec:compression_and_code_intelligence}
We investigate the relationship between compression ratio and code intelligence. 
We first propose the sliding window BPC for quantifying compression ratio and construct a set of robust validation corpus. Then, 
we present OmniCodeBench, a multi-lingual and multi-task code benchmark for comprehensive evaluation of code intelligence capabilities of LLMs.

\subsection{Compression Ratio and Validation Corpora}
\label{sec:sec:compression}
\subsubsection{Compression Ratio Criteria}
The paradigm of evaluating LLM intelligence through compression ratio necessitates a quantifiable criteria.
Bits per character (BPC) is a well-established metric for text compression, measuring the degree to which a LLM compresses a given text.
Formally, BPC can be defined as follows:
\begin{equation}
\begin{aligned}
BPC = \sum_{i=1}^N \log\left(\frac{1}{p_i}\right) \cdot \frac{1}{M} 
= \underbrace{\frac{1}{N} \sum_{i=1}^N \left(-\log(p_i)\right)}_{\mathrm{train ~ loss}} \cdot \frac{N}{M}
\\
= R_1 \cdot R_2 = R
\end{aligned}
\end{equation}
where $M$, $N$ denote corpus size and vocabulary size, respectively.
$R_1$, the first term of the BPC formula, captures model compression through next-token prediction loss, which is equivalent to the training loss upon first exposure.
$R_2$, the second term of the BPC formula,  represents vocabulary compression rate.
It is worth noting that due to differences in vocabulary size, cross-family model comparisons need to consider $R_1 \cdot R_2$, while analysis within the same family can focus solely on $R_1$.


\subsubsection{Sliding Window BPC}

A critical yet frequently overlooked factor in vanilla BPC evaluation lies in the heterogeneity of models' context window lengths $W$. 
Previous practices often simply truncate validation samples exceeding $W$, thereby discarding cross-chunk contextual dependencies and introducing bias when comparing models with divergent context length.
To address this, we propose a sliding window protocol with stride adaptation for precise BPC calculation.
For each input data sequence, we employ a sliding window with window size W and stride W/4 to compute logprobs. With the exception of fully retaining logprobs from the initial window, subsequent computations preserve only the non-overlapping segment at the window's trailing edge (length = stride size W/4). This strategic retention mechanism approximates contextual relationships while achieving computational efficiency equivalent to 7/8 of the full context length utilization of the original model.

The adjusted sliding window BPC metric is formalized as:
\begin{equation}
\text{BPC}_{\text{aj}} = \frac{1}{M} \sum_{k=0}^{K} \sum_{i=kW}^{(k+1)W} \mathbb{I}_{\text{valid}(i)} -\log_2 p_\theta(x_i|x_{1:i-1})
\end{equation}
where $\mathbb{I}_{\text{valid}(i)}$ masks positions outside stride-aligned computation zones except for the initial window, and $V = W/4 $ ensures a moderate stride length.

\begin{figure*}
\begin{center}
   \includegraphics[width=0.8\linewidth]{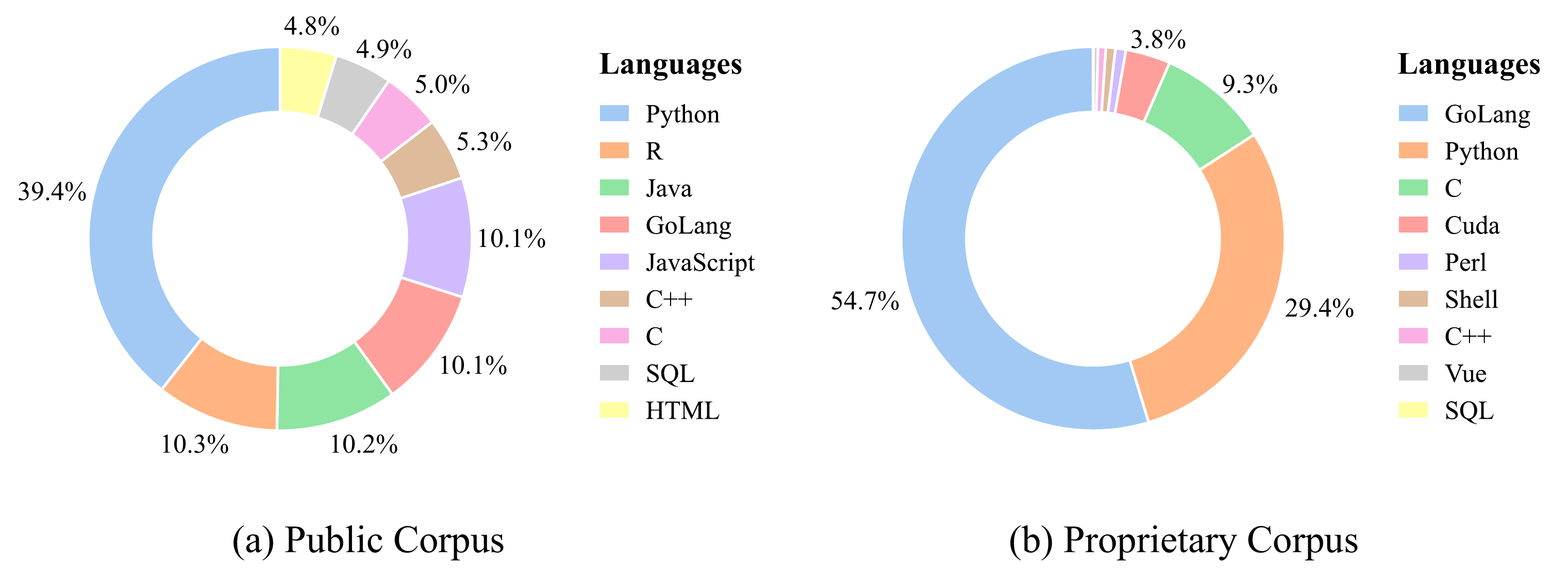}
\end{center}
   \caption{Programming language token distribution in the open-source and proprietary validation set. In (b), the unlabeled categories Perl, Shell, C++, Vue, and SQL account for 0.88\%, 0.80\%, 0.66\%, 0.38\%, and 0.03\%, respectively.}
\label{fig:valset}
\end{figure*}

\subsubsection{Robust Validation Corpus Construction}
As a proxy metric, the reliability of BPC to accurately reflects code intelligence heavily relies on the quality of the validation set.
To obtain a robust and high-quality validation set, we construct the validation corpus based on the following fundamental principles:
(1) Unseen integrity: All validation samples are rigorously excluded from the training data of any model. 
(2) Bias-free composition: Strict prohibition of model-generated or model-processed content to prevent overfitting towards specific model families. 
(3) Diversity: Ensuring linguistic coverage of mainstream programming languages through extensive collection and randomized sampling protocols. 
(4) High quality: Validation samples must maintain strict syntactic correctness and logical integrity to avoid distortion of compression rate measurements. Low-quality code containing redundant noise (e.g., duplicate code blocks), syntax errors, or incomplete semantic structures would artificially inflate BPC values due to invalid token patterns, thereby undermining the metric's ability to reflect true model capabilities.

Following the aforementioned criteria, we construct two high-quality corpora, one based on proprietary internal data and the other on publicly available open-source data.

\textbf{Proprietary corpus}~~
We adopt high-quality internal repository data as the source to construct a proprietary validation set. 
Our filtering pipeline proceeds as follows:
First, we employ a customized tokenizer to remove files shorter than 128 tokens, ensuring only sufficiently large code blocks are retained.
Subsequently, we filter out data that does not contribute meaningfully to the code content, such as copyright headers and metadata.
To further enhance quality, we apply a heuristic strategy to exclude repositories containing excessive prompt-engineered synthetic data, thereby preserving the authenticity of the code, following by LLM-based syntax check.
Finally, a thorough manual review is conducted by a team of three senior developers to identify any potential issues or inconsistencies.
This multi-layered approach ensures that the final proprietary corpus is both high-quality and representative of real-world software development practices.




\textbf{Public corpus}~~
To mitigate linguistic diversity limitation and distributional bias in proprietary datasets, we additionally construct an open-source validation set.
We select GitHub as our data source and employ timestamp-based filtering to prevent validation data leakage.
Specifically, based on the cutoff date of existing code LLM pre-training data, we collect all GitHub repositories created between May 2024 and November 2024. Subsequently, we follow OpenCoder's~\cite{huang2024opencoder} pipeline to perform filtration, removing non-code data and deduplicating the dataset. 
To minimize bias, we employ rule-based filtering rather than model-based filtering approaches.
Additionally, we parse the code to confirm its compilability and apply heuristic-based selection (e.g., filtering by natural language comment length) to preserve high-quality data. 
However, due to potential historical code leakage from GitHub branches, we adopt MinHash-LSH algorithm to prevent the inclusion of duplicate or near-duplicate code in the validation set.
To ensure balanced domain and language distribution in the development set, we perform weighted sampling by repository category, mitigating bias toward high-resource languages.
The distribution of validation set is illustrated in Figure~\ref{fig:valset}.



\subsection{Precise Evaluation of Code Intelligence: Benchmark and Technique}
\label{sec:sec:code_intelligence}
\subsubsection{OmniCodeBench}
The evaluation of code intelligence in LLMs remains an open research question. 
While existing studies predominantly rely on Python-specific benchmarks like HumanEval~\cite{chen2021evaluating} and MBPP~\cite{austin2021program}, Python alone cannot represent the full spectrum of programming concepts~\cite{luo2024python}. 
Our evaluation, therefore, encompasses \textbf{multiple programming languages}.
Moreover, conventional assessments focus primarily on the code completion task, which constitutes merely a fraction of real-world coding intelligence.
We address this limitation through a comprehensive \textbf{multi-task} evaluation framework. 
To this end, we propose \textbf{OmniCodeBench}, a \textbf{multi-lingual} and \textbf{multi-task} code benchmark capable of holistic and robustly evaluating code intelligence.
Concretely, \textbf{OmniCodeBench} consists of McEval~\cite{chai2024mceval} for multi-lingual code generation, BIRD~\cite{li2023can} and Spider~\cite{yu2018spider} for SQL code generation,  LiveCodeBench~\cite{jain2024livecodebench} for program repair, HumanEvalExplain~\cite{muennighoff2023octopack} for code explanation, CRUXEval~\cite{gu2024cruxeval} for code reasoning, and HumanEval~\cite{chen2021evaluating}, MBPP~\cite{austin2021program}, BigCodeBench~\cite{zhuo2024bigcodebench}.

\begin{table}
\caption{Comparison of the number of empty strings before and after format annealing for five models. The empty strings are counted from the preprocessed responses on the MBPP+ benchmark.}
\centering
\scalebox{0.9}{
\begin{tabular}{lcc}

\toprule
\textbf{Model} & \textbf{Before annealing} & \textbf{After annealing} \\
\midrule
StarCoder2-3B & 136 & 1 \\
CodeGemma-2B & 55 & 2 \\
CodeLlama-7B & 10 & 2 \\
Qwen2.5-Coder-7B & 13 & 4 \\
Granite-8B-Code & 12 & 3 \\
\bottomrule
\end{tabular}}
\label{tab:empty_strings}
\end{table}
\subsubsection{Format Annealing}

Code-related tasks require extracting code from the model's responses and verifying its correctness through test cases.
However, the response format of the model is often difficult to perfectly cover, leading to omissions during extraction.
Additionally, the model frequently outputs meaningless placeholders, such as \texttt{todo} and \texttt{pass}, severely impacting the accurate evaluation of code intelligence. This phenomenon is caused by the excessive occurrence of related patterns in the pre-training data~\cite{huang2024opencoder}.
To address this, we construct an annealed training set to ensure: (1) preventing external knowledge infusion beyond the original capacity of foundation models, and (2) maintaining strict format compliance with downstream benchmarks.
Following these requirements, we integrate three data sources: CommitPackFT~\cite{muennighoff2023octopack}, CodeSearchNet~\cite{husain2019codesearchnet}, and LeetCode~\cite{huang2024opencoder}. 
All three datasets originate from GitHub, which is the major component of the pretraining corpora of Code LLMs. 
We run the data cleaning pipelines of various Code LLMs on this data, ensuring the remaining parts were seen during pre-training.
Subsequently, all non-functional placeholders are removed, and the content is reformatted according to the prompts of downstream benchmarks, ensuring alignment for evaluation.
The final curated dataset contains 4,995 training samples. 
The complete data synthesis pipeline for each dataset is presented in the supplementary material.

The curated dataset is then used for the specialized continual training phase for all models. During this phase, after each epoch, we leverage the MBPP+~\cite{evalplus} to assess the proportion of post-processed outputs that are empty strings.
Training for each model is terminated when this proportion drops to 1\%. This early stopping strategy is employed to minimize alterations to the models' original parameter distributions. 
Table~\ref{tab:empty_strings} shows the comparison of the number of empty strings before and after format annealing for five models. 
It shows that format annealing effectively mitigates the inherent limitations of foundational models in format compliance, ensuring fair comparison through format-controlled evaluation.

\section{Experiments}
\subsection{Setup}
\paragraph{Models}
Multiple state-of-the-art Code LLMs are adopted, covering model scales from 1.5B to 70B parameters. 
Selected models are listed as follows:
Qwen2.5-Coder~\cite{hui2024qwen2}, 
DeepSeek-Coder~\cite{guo2024deepseek}, 
StarCoder~\cite{li2023starcoder, lozhkov2024starcoder}, 
CodeGemma~\cite{team2024codegemma}, 
CodeLlama~\cite{roziere2023code}, 
Llama3.1~\cite{grattafiori2024llama},
Yi-Coder~\cite{yicoder}, Granite-Code~\cite{mishra2024granite},
and OpenCoder~\cite{huang2024opencoder}.

\paragraph{Hyperparameters}
Our format annealing training set consists of 4,995 QA pairs. 
We use DeepSpeed ZeRO-3 optimization with BF16 precision on 8*A100-80GB GPU.
The batch size, learning rate~\cite{li2025predictablescalei}, and warm-up ratio are set to 128, 1e-4, and 0.1, respectively.
Each models are trained for up to 4 epochs. After each epoch, the ratio of successful extraction from outputs is evaluated. When it exceeds 99\%, the training is stopped and the current checkpoint is used for the OmniCodeBench evaluation.

For all benchmark evaluations, we employ consistent configurations: a temperature of 0.2, a top-p value of 0.95, and 10 samples generated per query. 
The average pass rate of these 10 samples is used as the final result.

\paragraph{Metrics}
Our metric framework uses a two-stage weighting approach. First, tasks across code domains (e.g., code fix, generation) are weighted uniformly. Second, intra-task weights are scaled based on the number of test instances in each benchmark. The composite capability score \( C \) is computed as follows:
\begin{equation}
    C = \frac{\sum_{k=1}^N \sum_{j=1}^{M_k} w_{k,j} \cdot s_{k,j}}{\sum_{k=1}^N \sum_{j=1}^{M_k} w_{k,j}}
\end{equation}
\( N \) denotes the number of code task categories, \( M_k \) represents the benchmarks in task \( k \), \( w_{k,j} = \frac{1}{N} \cdot \frac{|D_{k,j}|}{\sum_{i=1}^{M_k} |D_{k,i}|} \) is the hierarchical weight for benchmark \( j \) in task \( k \), and \( s_{k,j} \in [0,1] \) is the performance on benchmark \( j \).This entropy-based weighting strategy ensures initial task fairness via the \( \frac{1}{N} \) term and a data-driven benchmark emphasis via the \( |D_{k,j}| \) proportion.
After extensive experiments comparing the fitting curves of ${C}$ and $\log{C}$, We choose $\log{C}$ as our final metric, which represents code intelligence. Relevant experiments are presented in the supplementary material.
    
\subsection{Results and Findings}



\begin{figure*}
\begin{center}
   \includegraphics[width=0.9\linewidth]{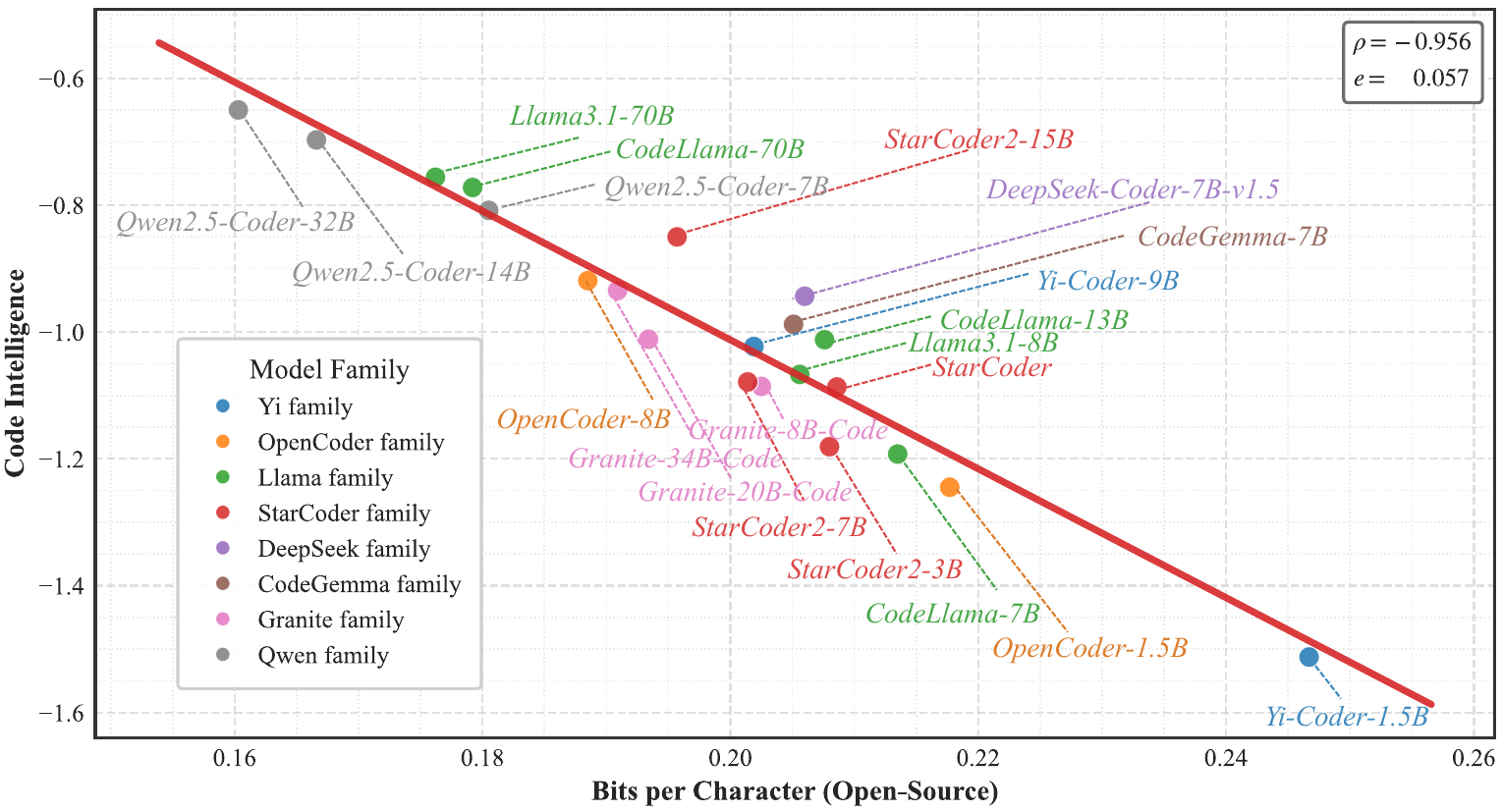}
\end{center}
   \caption{Logarithmic relationship between BPC and code intelligence on our open-source validation set. The values of code intelligence are pre-processed in logarithmic scale from benchmark metrics covering multiple tasks and languages.}
\label{fig:bpc_all_score_all}
\end{figure*}
\begin{figure*}
\begin{center}
   \includegraphics[width=0.9\linewidth]{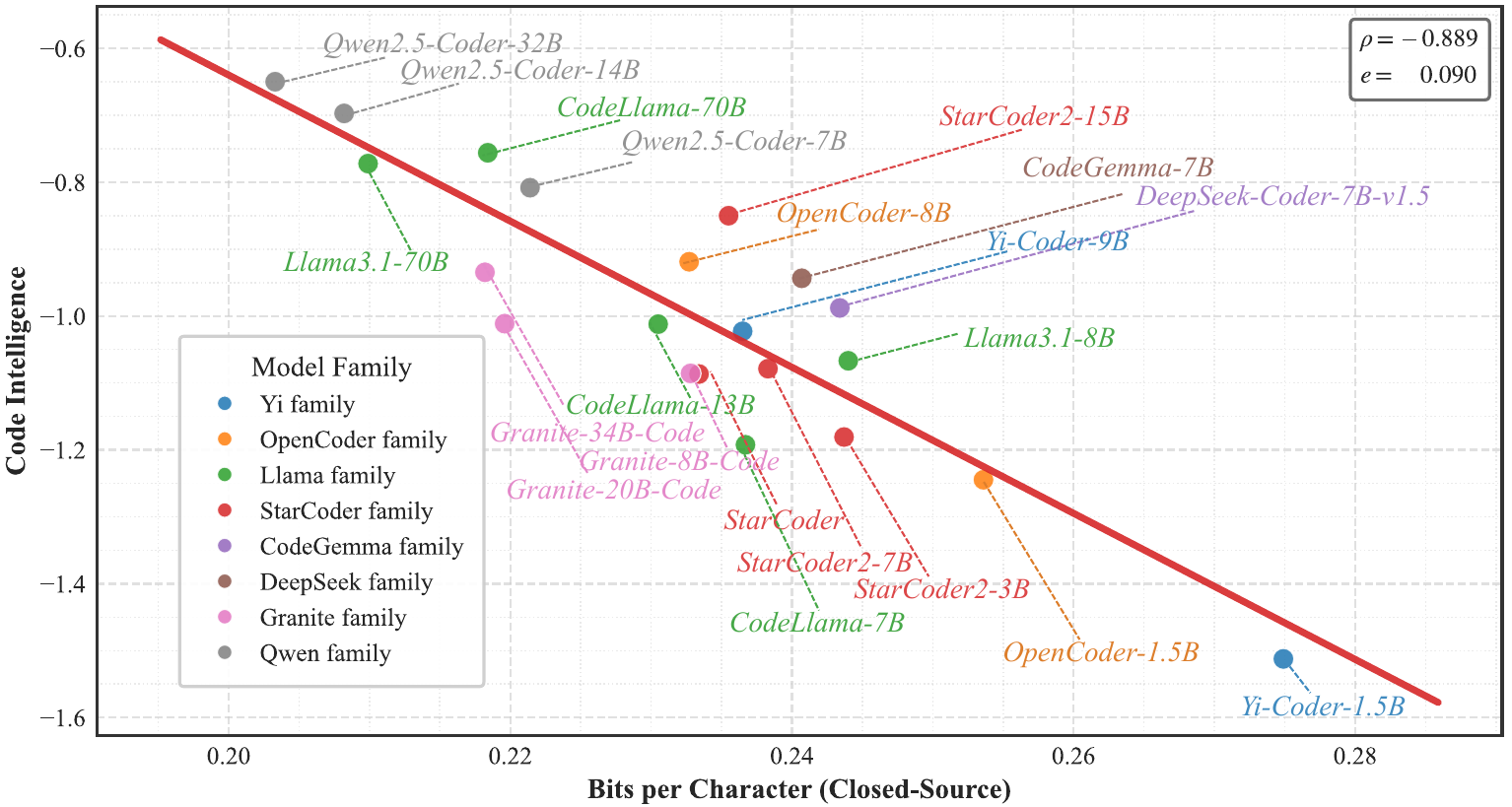}
\end{center}
   \caption{Logarithmic relationship between BPC and code intelligence on our closed-source validation set. The values of code intelligence are pre-processed in logarithmic scale.}
\label{fig:bpc_all_score_all_close}
\end{figure*}



Figure~\ref{fig:bpc_all_score_all} reveals an approximately linear correlation between code BPC on our open-source validation set and the log-transformed code intelligence metric with a Pearson coefficient of -0.956 and a RMSE (Root Mean Square Error) of 5.7\%, while Figure~\ref{fig:bpc_all_score_all_close} shows the same curve fitting correlation on our closed-source validation set with a Pearson coefficient of -0.956 and a RMSE of 9\%. This performance discrepancy can be attributed to more balanced data distribution and stricter anti-leakage design in the open-source validation set, which align with real-world code distribution patterns better than the closed-source validation set.
The results suggest a \textbf{logarithmic} rather than a linear relationship between code BPC and code intelligence contrary to the assumptions in prior works. 
    

We think that this logarithmic relationship aligns more in line with theoretical expectations. From the perspective of model capabilities, when benchm  ark difficulty is appropriately moderate, models with high BPC lack sufficient knowledge to solve problems effectively, whereas those with relatively low BPC exhibit substantial evaluation metric improvements even with marginal BPC reductions. This nonlinear correspondence strongly rationalizes the observed logarithmic law.


Furthermore, Dual effects are observed through analyzing positional shifts of data points on the fitting curve before and after format annealing. 
On the one hand, for base models with relatively low code intelligence performance (e.g., CodeLlama-7B\&13B), their positions converge toward the fitted curve after significantly mitigating format errors in responses. 
On the other hand, those with high code intelligence also realign with the curve upon suppressing potential contamination from exposure to specific patterns or instructional data during pretrain.
These bidirectional corrections substantiate the effectiveness of format annealing in decoupling format compliance from code intelligence. Consequently, the methodology establishes a systematic evaluation framework that minimizes unfavorable factors between code intelligence and response format, which contribute to the robustness and fairness of code intelligence evaluations.

\section{Discussions}
\begin{figure}
\begin{center}
   \includegraphics[width=1.0\linewidth]{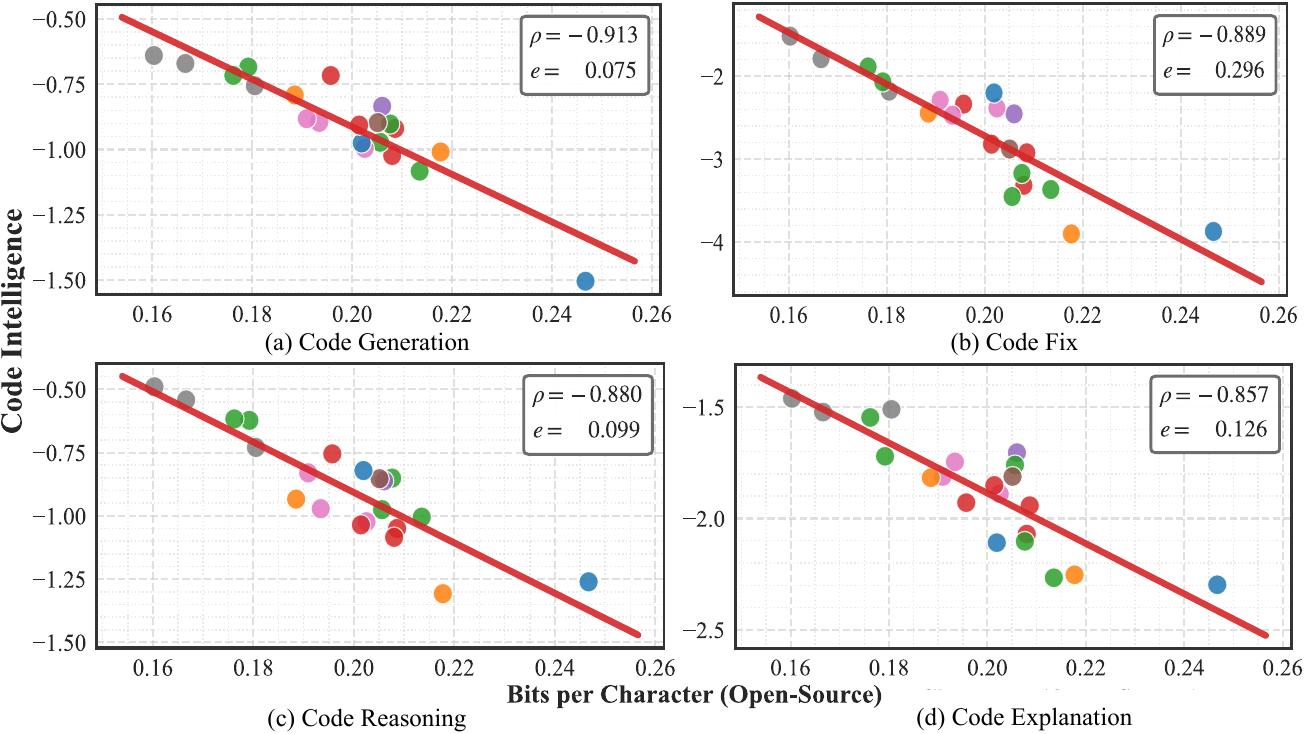}
\end{center}
   \caption{Fitting curves of four different code tasks (code generation, fix, reasoning and explanation). In each figure, benchmarks unrelated to the current task are discarded. The values of code intelligence are in logarithmic scale.}
\label{fig:four_task}
\end{figure}

\textbf{The necessity of multi-task: } Figure~\ref{fig:four_task} represents the relationship between BPC and code intelligence across four discrete tasks: code generation, fix, reasoning and explanation. The code fix task exhibit particularly high variance owe to its benchmark characteristics of limited and difficult test cases, while the code generation task appears to be relatively satisfactory with sufficient test cases. Overall, the task-specific fitting curve demonstrate obviously inferior to the multi-task one. This divergence validates the necessity of testing on multiple tasks, since the models perform unevenly across different tasks and focusing on any single one would lead to some bias. Therefore, multitask test is indispensable.

\begin{figure}
\begin{center}
   \includegraphics[width=1.0\linewidth]{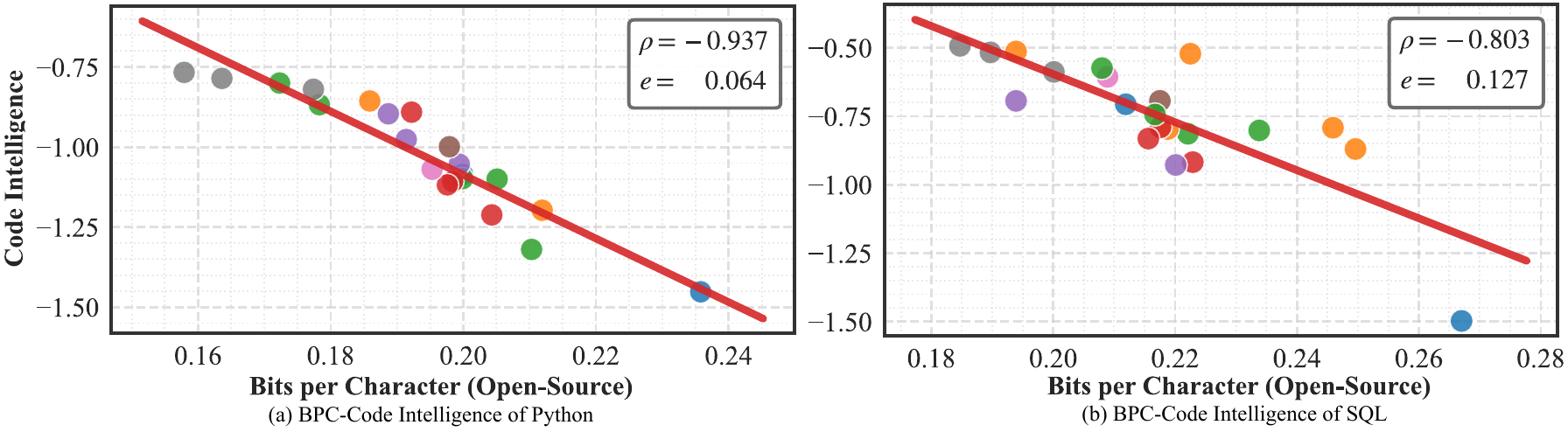}
\end{center}
   \caption{Fitting curves of two different programming languages (python, sql). In each figure, benchmarks unrelated to the current programming language are discarded, and BPC is calculated only on this language. The values of code intelligence are pre-processed in logarithmic scale.}
\label{fig:two_language}
\end{figure}

\textbf{The necessity of multi-language: }Similarly, we seperate specific programming languages to examine their relationships between BPC and code intelligence. Figure~\ref{fig:two_language} compares the two most tested languages Python and SQL in our benchmarks. The results show that Python performs well in curve fitting while SQL exhibits poorer correlation. The disparity mainly stems from the Python centric pretrain data of the models resulting in inconsistent SQL capabilities. Therefore, multilingual evaluation minimally impacts Python conclusions but enables broader generalization for overall coding competence of models.

\textbf{Abnormal impact from benchmarks: }During a single benchmark fitting process, we observe that McEval struggles to establish logarithmic relationships. Further analysis of the outputs reveal that on one hand, models with abnormally low McEval scores consistently produce lengthy invalid continuations containing various repetitions, meaningless characters, or multilingual noise. This issue appears most severe in Qwen-series models, causing their scores to drop to single-digit levels. On the other hand, models with unusually high McEval scores are disproportionately boosted by their C/C++ performance. For instance, OpenCoder-1.5B significantly outperforms OpenCoder-8B solely due to its superior C/C++ metrics, while both models exhibit normal Python performance. The inherent difficulty and linguistic diversity of McEval make it challenging to serve as a standalone reference point for observing our fitting logarithmic relationship.

\textbf{The profound significance of BPC: }Our findings offer BPC as a novel standard for code intelligence estimation. Instead of heavy traditional evaluation work, BPC only need a high-quality validation set having no intersection with pretrain data of the models. It is simpler to maintain the validation set than constantly updating new benchmarks. Although the current fitting accuracy is not sufficient to support accurately calculating code intelligence by BPC or any certain benchmark metric, BPC demonstrates enormous research potential and application prospects.

\textbf{Insight for further research of LLM: }Based on and inspired by our conclusions, this approach enables more precise modeling of the relationship between model compression and intelligence in code-related domains and even across general fields. This advancement not only facilitates rapid performance estimation for smaller models but also allows predictive scaling for larger-parameter models, thereby avoiding substantial redundant computational expenditure.

\section{Conclusion}    
This paper explores the logarithmic relationship between code BPC and code intelligence. To this end,  
we introduce a sliding window evaluation technique to ensure fair comparisons across models with varying context window lengths, addressing potential biases in BPC calculation. To further strengthen our evaluation, we develop OmniCodeBench, a multilingual, multi-task benchmark, alongside a Format Annealing training methodology. This approach improves pattern adherence while preserving the intrinsic knowledge of foundational models, effectively minimizing format-related errors that commonly affect model outputs. Our findings provide new possibilities for investigating the relationship between compression and intelligence of LLMs.







 \bibliographystyle{scis}
 \bibliography{ref}

\begin{appendix}
\section{Data Synthesis Pipeline for Training Set}
The training set integrates three primary sources: CommitPackFT~\cite{muennighoff2023octopack}, CodeSearchNet~\cite{husain2019codesearchnet}, and LeetCode, each subjected to a specialized preprocessing workflow. The CommitPackFT data containing paired ``old contents'' and ``new contents'' code revisions serves as the foundation for the code fix task. From an initial pool of approximately 50,000 sampled instances, we implemented a multistage cleaning protocol: first removing entries where old and new contents are identical or contain empty inputs, followed by filtering out files with non-executable placeholders (e.g., ``TODO'', ``FIXME''). We subsequently leveraged GPT-4o to automatically verify that the code in old content reliably triggers run-time errors while their corresponding new content resolve these issues, preserving associated error logs for prompt generation. A final manual inspection phase eliminates ambiguous or meaningless cases, resulting in 3,855 rigorously validated samples. These curated instances are further standardized using GPT-4o to match the input-output formatting specifications of LiveCodeBench code fix benchmark, ensuring seamless integration with downstream evaluation frameworks.

The LeetCode dataset comprises 3,077 programming challenges and their corresponding solutions scraped from LeetCode international platform, exhibiting inherently high data quality due to its curated nature. To clean the data, we implemented a standardized filtering protocol: (1) excluding video-based solutions; (2) randomly selecting a solution for each problem; (3) structurally aligning each instance with LiveCodeBench's code generation task format with the assistance of GPT-4o. The standardized prompt template integrates four key components: the problem statement, illustrative examples (including input and output cases), algorithmic constraints and the initial function signature, thereby preserving the essential context required for reproducible code generation tasks.

The CodeSearchNet dataset extracts function docstrings as instructions, with a high degree of structural repetition. We removed entries where the output character length exceeds 5,000, as well as those with too short or meaningless instructions. The data was then sorted in descending order by length, and 5-gram deduplication was applied to removing any instructions or output containing ``TODO'' or ``FIXME''. We used GPT-4o to filter suitable data for each language, and finally, a comprehensive sampling was performed to obtain 8,699 data samples.
Ultimately, we sampled 4,995 training data samples from three data sources as our training set.

\section{Why Choose LogC}

\begin{figure*}
\begin{center}
   \includegraphics[width=0.8\linewidth]{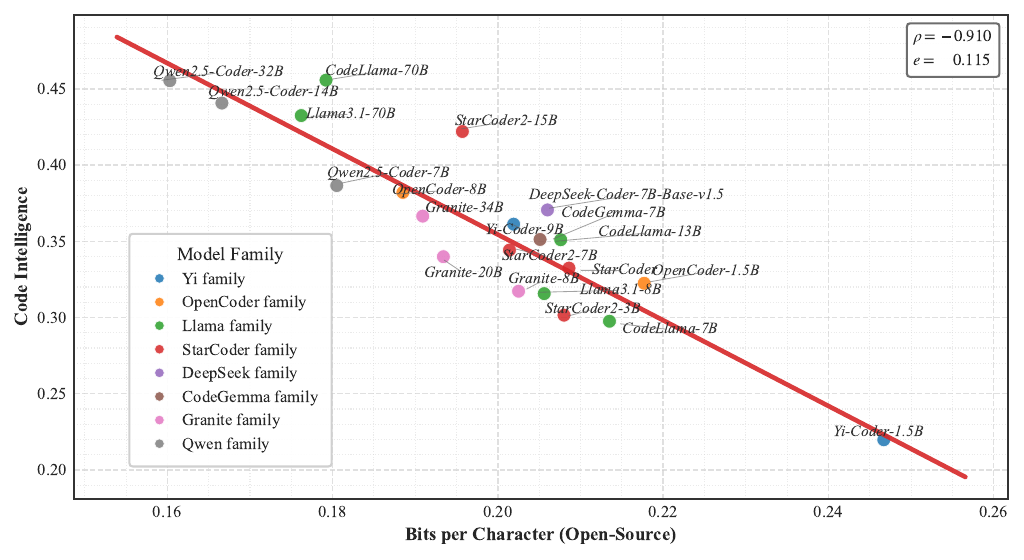}
\end{center}
   \caption{Linear relationship between BPC and code intelligence on our open-source validation set.}
\label{fig:bpc_all_score_all_linear}
\end{figure*}

Figure~\ref{fig:bpc_all_score_all_linear} shows that with a RMSE of 0.115 the linear fitting curve performs worse than the logarithmic relationship curve with a RMSE of only 0.057. Therefore, in our paper the values of code intelligence are all pre-processed in logarithmic scale.

\section{Individual Fitting Curve of Each Benchmark}
\begin{figure*}
\begin{center}
   \includegraphics[width=0.8\linewidth]{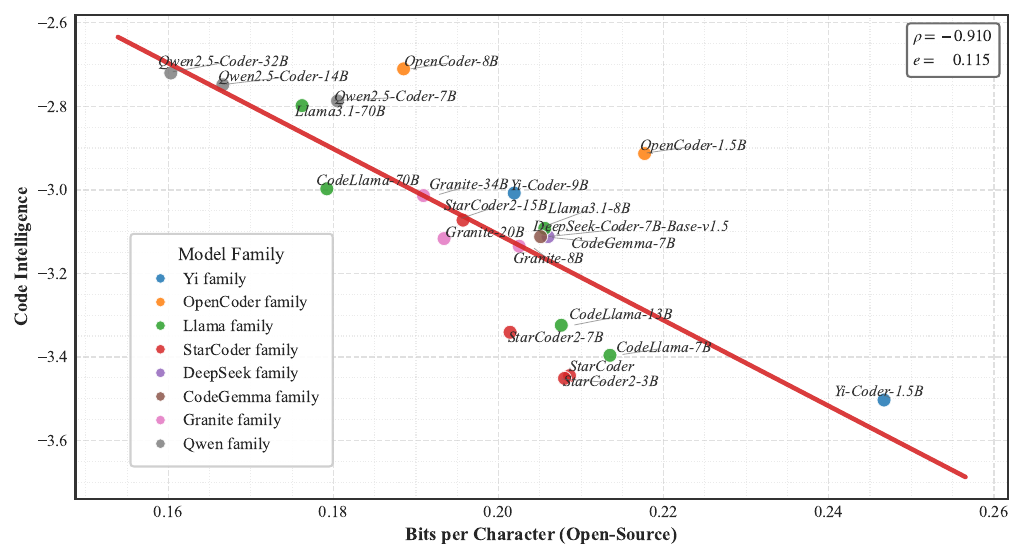}
\end{center}
   \caption{Fitting curve between HumanEval benchmark and its corresponding metric.}
\label{fig:1}
\end{figure*}

\begin{figure*}
\begin{center}
   \includegraphics[width=1.0\linewidth]{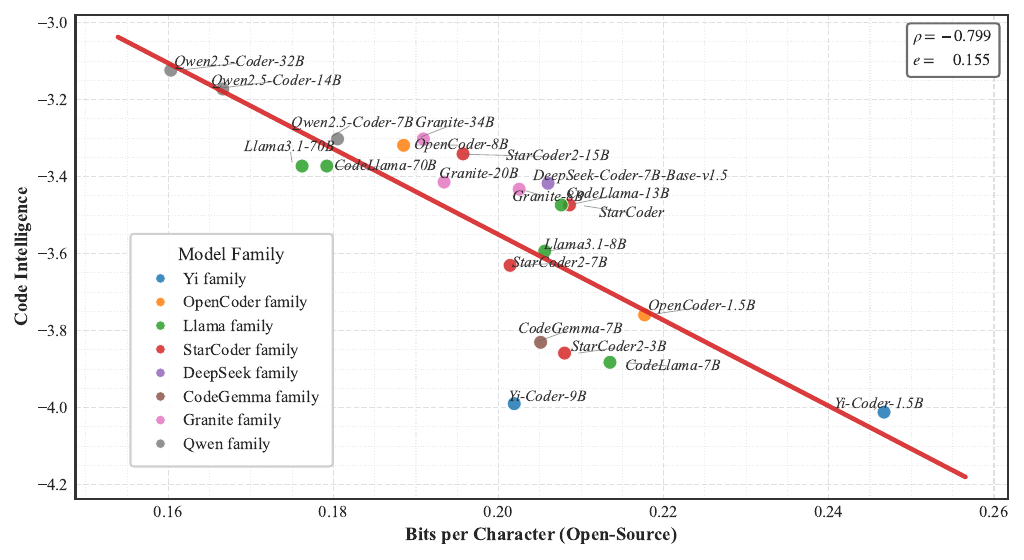}
\end{center}
   \caption{Fitting curve between BigCodeBench-Full benchmark and its corresponding metric.}
\label{fig:4}
\end{figure*}

\begin{figure*}
\begin{center}
   \includegraphics[width=1.0\linewidth]{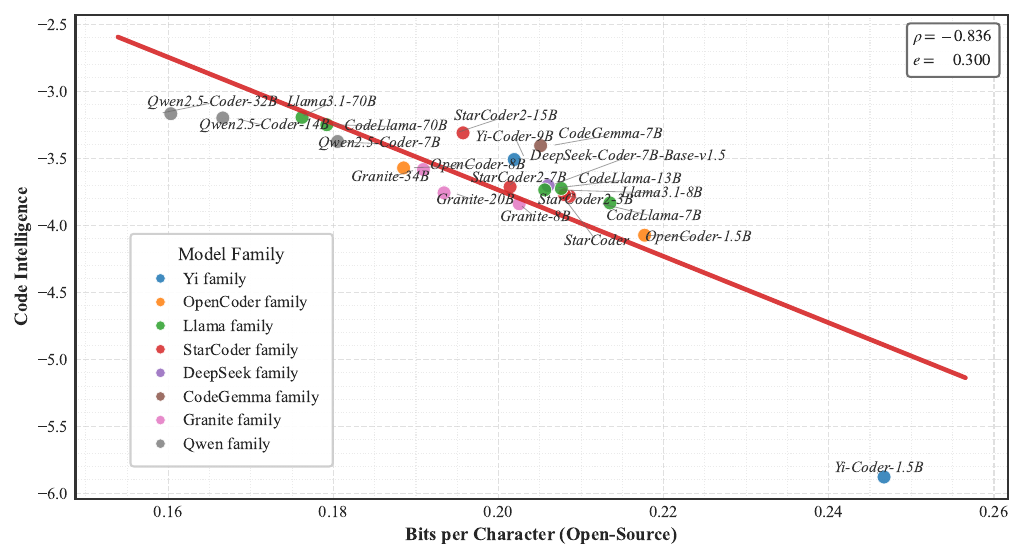}
\end{center}
   \caption{Fitting curve between BIRD benchmark and its corresponding metric.}
\label{fig:5}
\end{figure*}

\begin{figure*}
\begin{center}
   \includegraphics[width=1.0\linewidth]{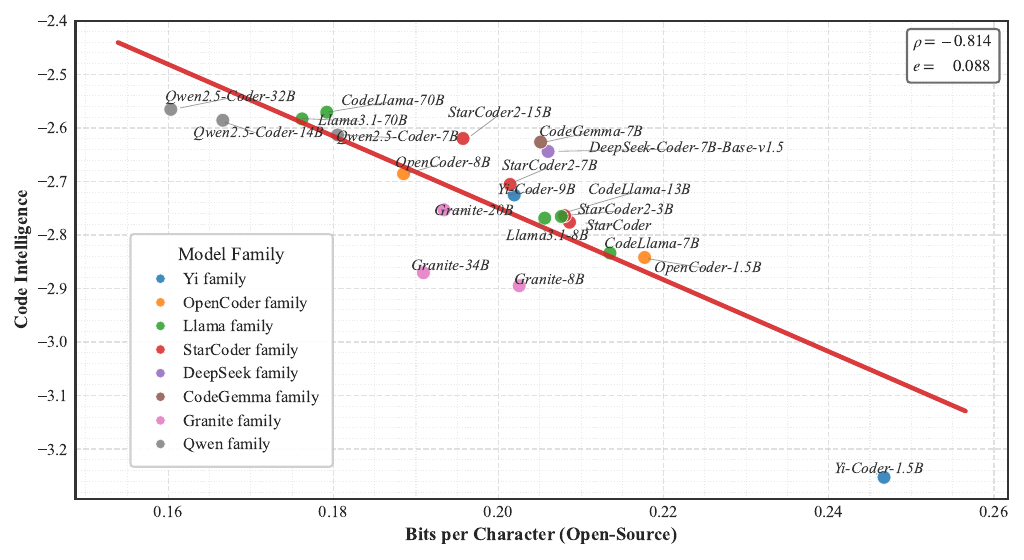}
\end{center}
   \caption{Fitting curve between Spider benchmark and its corresponding metric.}
\label{fig:6}
\end{figure*}

\begin{figure*}
\begin{center}
   \includegraphics[width=1.0\linewidth]{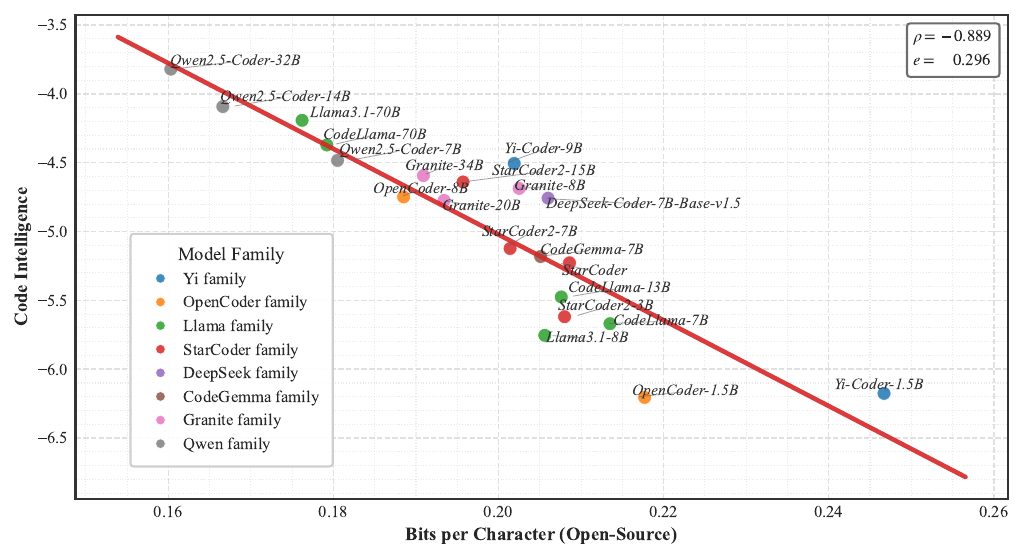}
\end{center}
   \caption{Fitting curve between LiveCodeBench-Fix benchmark and its corresponding metric.}
\label{fig:7}
\end{figure*}

\begin{figure*}
\begin{center}
   \includegraphics[width=1.0\linewidth]{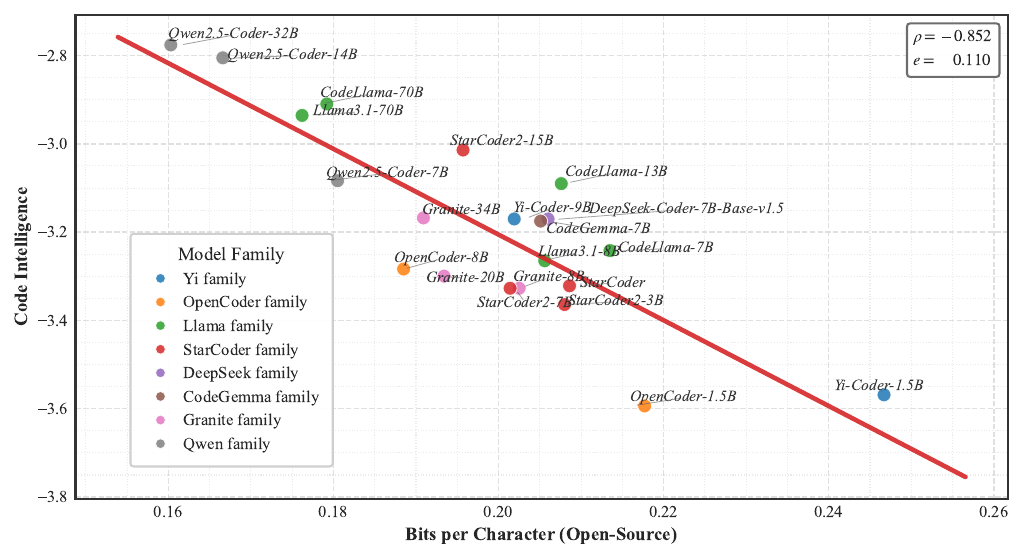}
\end{center}
   \caption{Fitting curve between CRUXEval-I benchmark and its corresponding metric.}
\label{fig:8}
\end{figure*}

\begin{figure*}
\begin{center}
   \includegraphics[width=1.0\linewidth]{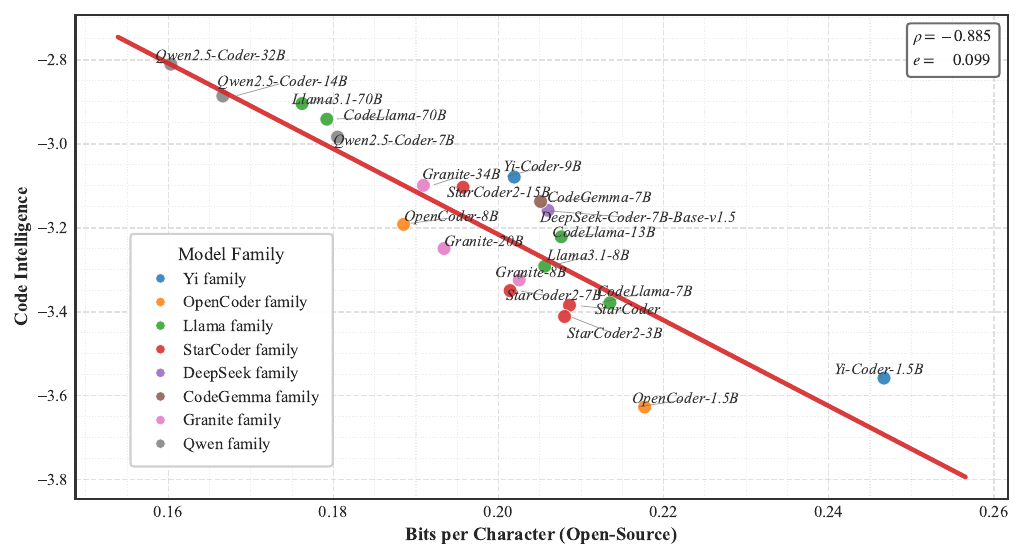}
\end{center}
   \caption{Fitting curve between CRUXEval-O benchmark and its corresponding metric.}
\label{fig:9}
\end{figure*}

\begin{figure*}
\begin{center}
   \includegraphics[width=1.0\linewidth]{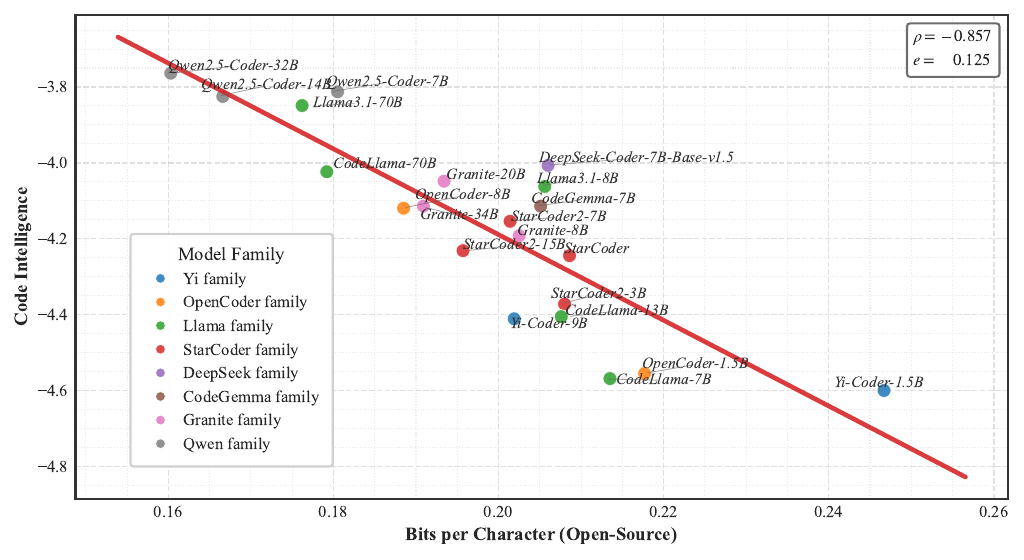}
\end{center}
   \caption{Fitting curve between HumanEvalExplain benchmark and its corresponding metric.}
\label{fig:10}
\end{figure*}

 The fitting results of each benchmark are displayed in Figure~\ref{fig:1}, ~\ref{fig:2}, ~\ref{fig:3}, ~\ref{fig:4}, ~\ref{fig:5}, ~\ref{fig:6}, ~\ref{fig:7}, ~\ref{fig:8}, ~\ref{fig:9}, ~\ref{fig:10} which seperately represent HumanEval, BigCodeBench-Full, BIRD, Spider, LiveCodeBench-Fix, CRUXEval-I, CRUXEval-O and HumanEvalExplain.



\newpage
\section{Details of Open-Source Validation Set}
 Detailed data quantity and proportion of our open-source validation set are shown in Table~\ref{tab:valset}.

\begin{table}
\caption{More details of our open-source validation set.}
\centering
\begin{tabular}{lccc}
\toprule
Language   & Number of Data Entries & Number of Characters & Proportion of Characters \\
\midrule
Python     & 1, 500                 & 7, 086, 432     & 41.21\%             \\
Go         & 500                    & 1, 598, 829        & 9.30\%             \\
R          & 300                    & 1, 460, 408        & 8.49\%             \\
C          & 100                    & 714, 577        & 4.16\%              \\
C++        & 200                    & 861, 125        & 5.01\%              \\
Java       & 500                    & 2, 020, 641        & 11.75\%             \\
JavaScript & 500                    & 1, 866, 524        & 10.85\%             \\
HTML       & 100                    & 892, 826        & 5.19\%              \\
SQL        & 200                    & 695, 673        & 4.05\%              \\
Total      & 3, 900                 & 17, 197, 035     & 100.00\%  \\  
\bottomrule
\end{tabular}
\label{tab:valset}
\end{table}

\end{appendix}

\end{document}